\pdfoutput=1
\documentclass[runningheads]{llncs}

\usepackage[pdftex]{graphicx}
\usepackage{amsmath}
\usepackage{threeparttable}
\usepackage{xcolor}
\usepackage{hyperref}
\newcommand\blfootnote[1]{%
    \begingroup
    \renewcommand\thefootnote{}\footnote{#1}%
    \addtocounter{footnote}{-1}%
    \endgroup
}

\begin{document}
\title{Cascaded Classifier for Pareto-Optimal Accuracy-Cost Trade-Off Using off-the-Shelf ANNs}
\titlerunning{Cascaded Classifier for Pareto-Optimal Accuracy-Cost Trade-Off}

\author{Cecilia Latotzke\orcidID{0000-0001-6536-820X} \and
Johnson Loh\orcidID{0000-0002-0966-8918} \and
Tobias Gemmeke\orcidID{0000-0003-1583-3411}}

\authorrunning{C. Latotzke et al.}

\institute{RWTH Aachen University, Aachen 52062, Germany\\
\email{\{latotzke,loh,gemmeke\}@ids.rwth-aachen.de}
\blfootnote{The authors C. Latotzke and J. Loh contribute equally to the paper.
The final accepted publication was presented on the 7th International Conference on Machine Learning, Optimization, Data Science (LOD), October 4 – 8, 2021 in Grasmere, Lake District, England.}
}

\maketitle
\begin{abstract}
Machine-learning classifiers provide high quality of service in classification tasks.
Research now targets cost reduction measured in terms of average processing time or energy per solution. Revisiting the concept of cascaded classifiers, we present a first of its kind analysis of optimal pass-on criteria between the classifier stages. Based on this analysis, we derive a methodology to maximize accuracy and efficiency of cascaded classifiers. 
On the one hand, our methodology allows cost reduction of 1.32$\times$ while preserving reference classifier’s accuracy. On the other hand, it allows to scale cost over two orders while gracefully degrading accuracy.
Thereby, the final classifier stage sets the top accuracy. Hence, the multi-stage realization can be employed to optimize any state-of-the-art classifier.
\keywords{Cascaded classifier \and machine learning \and edge devices \and preliminary classifier \and pareto analysis \and design methodology.}
\end{abstract}

\section{Introduction}
Machine learning techniques are on the rise for classification tasks since they achieve higher accuracies than traditional classification algorithms. Especially, Deep Neural Networks (DNN) have proven highly effective in benchmark competitions such as the ImageNet Challenge \cite{ImageNet}. 
However, this success comes at a price - computational complexity and thereby energy has skyrocketed \cite{Canziani2017}.

Today, automated data analysis is used in sensitive medical or industrial applications.
These tasks vary over a wide range such as image classification, voice recognition or medical diagnostic support. 
As misclassification in these areas has a high price, classification accuracy is expected to be on par with a trained human. 

In some application cases, raw data contains sensitive private information, i.e.,  streaming to and processing in the cloud is not always an option, letting alone the energy cost of transferring raw data as compared to classification results.
This together gives rise to classification on edge devices.
Featuring a limited energy budget, the computational cost of classification is constrained \cite{Li2009}.
Preferably, the classification algorithm needs to be adjustable to balance between the available energy budget and application specific accuracy requirements.
Furthermore, operating in real-time drives the need for high throughput. At the same time, live data contains identical or even irrelevant information, i.e., the data-sets are skewed \cite{Cocana2017,Goetschalckx2018}.
Thus, a promising approach for cost reduction is preprocessing in wake-up or reduced-complexity classifiers that forward only seemingly interesting samples towards the next stage.
This was successfully applied in embedded sensor nodes \cite{Benbasat2007} and for speech recognition \cite{Badami2015,Price2017}. 
The goal is to achieve lowest possible energy for data processing on the edge device while maintaining a high quality of service of the classification algorithm \cite{Latotzke2021}. 
They can be summarized as hierarchical classifiers.

The general concept of hierarchical classifiers was first introduced by Ouali and King \cite{Ouali2000}.
The various design dimensions such as applied algorithm, architecture, or quantization span a large design space for cascaded classifiers. Adding the substantial variety of existing datasets and their complex feature space, benchmarking different solutions becomes a challenge by itself \cite{GoensCastrillon2019}.

This work focuses on the optimization of cascaded classifiers to accelerate highly energy efficient classification on edge devices.
The main contributions of this manuscript are:
1) a first of its kind in-depth analysis of various existing and novel pass-on criteria between the stages of a cascaded classifier, used to qualify a classifier's output and steer the pass-on-rate, the rate of samples not classified in the current classifier stage and passed to the succeeding classifier stage;
2) a detailed discussion of different design choices of cascaded classifiers as low-cost in-place substitutes of an existing classifier;
3) a design methodology based on the accuracy-cost trade-off for Pareto-optimal cascaded classifier constellations, with cost in terms of average processing time or energy per sample
and 4) the composition of accuracy-efficiency trade-off driven cascaded classifiers, of in-house trained classifiers for both datasets, MNIST \cite{Lecun2010} and CIFAR \cite{Krizhevsky2009}.
Their Pareto-optimal settings provide a graceful trade-off between classification accuracy and cost per classification -- matching the specific use case. 
We provide the template scripts to reproduce the results of this paper in \url{https://git.rwth-aachen.de/ids/public/cascaded-classifier-pareto-analysis}. 

The paper is organized as follows. 
Section 2 discusses the state-of-the-art of cascaded classifiers and introduces its design parameters. 
Based on the analysis of the pass-on criteria, the design methodology for an efficiency-accuracy driven cascaded classifier is presented in Section 3.
In Section 4 the validation of the methodology for CIFAR\thinspace{10} is shown. Conclusions and remarks are drawn in Section 5.

\section{Related work and background}

\begin{figure}[htbp]
	\centering
	\includegraphics[scale=0.9]{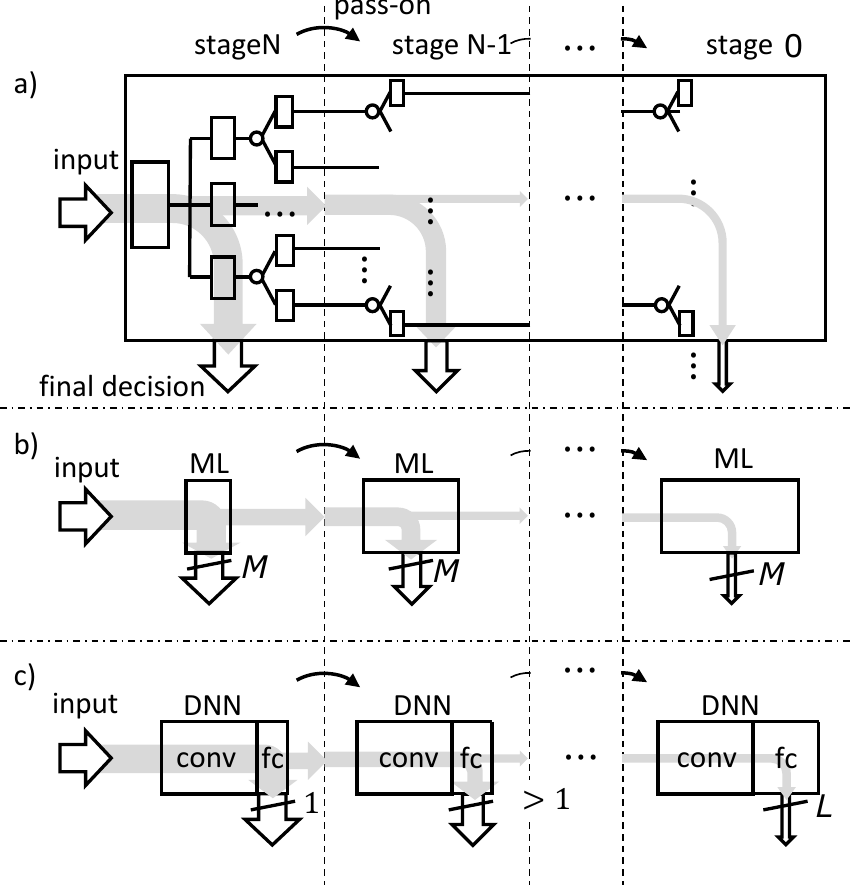}
	\caption{Concept of state-of-the-art cascaded classifiers with increasing (a) complexity of referenced features \cite{Cocana2017}, (b) quality of multi-class classifier \cite{Venkataramani2015}, \cite{Kouris2018} and (c) number of output classes \cite{Goetschalckx2018}, with M and L being the maximum number of classes for each model correspondingly.}
	\label{fig:s2_sota_cascaded_classifiers}
\end{figure}

The concept of hierarchical classification comprises various ways to combine classifiers including trees or cascades  \cite{Xu2014}.
This paper focuses without any loss of generality on the latter subdomain, i.\,e.\ classifier structures being activated sequentially.
The idea is to pass a sample over consecutive stages of increasing accuracy and cost until a stage's specific pass-on criterion is above a preset threshold.
The number of samples in the cascaded classifier is reduced by every decision made in a previous stage (grey arrows in Fig.\,\ref{fig:s2_sota_cascaded_classifiers}), while undecided samples are passed on to the next stage.
Per convention, the indexing starts with $i=0$ at the last stage - the reference classifier, that sets the maximum accuracy.

 An overview of state-of-the-art cascaded classifier architectures is shown in Fig.\thinspace\ref{fig:s2_sota_cascaded_classifiers}. The approach to develop classifier stages, pass samples and optimize the system varies in the different structures.
 
In Fig.\thinspace\ref{fig:s2_sota_cascaded_classifiers}a) the stages are using fuzzy trees to determine the classified label \cite{Cocana2017}. 
By increasing the number of features utilized by the decision tree, the cost is increased as well as its capability to achieve higher accuracy. The fuzzy classifier computes initially only a subset of all features, since not all samples are equally complex to classify. 
Additional and more complex features are added in later stages to enable proper classification of less frequent, more difficult to classify samples. 
Thereby, earlier extracted features are propagated resulting in a classifier with on-demand generation of computationally more expensive features \cite{Xu2013}. 
As a drawback, this multi-stage scheme is rather inflexible, as the architecture has to be revised for any change in the distribution of features contained in the samples.

A more general approach was earlier proposed in \cite{Venkataramani2015} (cf.\,Fig.\thinspace\ref{fig:s2_sota_cascaded_classifiers}b). 
The general concept is the concatenation of independent Machine Learning classifiers (ML), whose complexity is progressively increasing. It focuses on ensembles of binary ``one-class-expert'' classifiers determining the confidence value based on relative affiliation of a sample to a specific class.

Another principle is to modify the number of classes processed in each stage (cf.\,Fig.\thinspace\ref{fig:s2_sota_cascaded_classifiers}c{\small )}, \cite{Goetschalckx2018}). 
With increasing stage index the networks are trained to differentiate a larger number of classes in the set.
This approach can be especially beneficial in real-life scenarios with highly skewed input data, since many irrelevant, but frequent, samples are processed in the earlier stages.
Various examples exist that introduce such ''wake-up'' stage for energy optimization \cite{Badami2015,Benbasat2007,Price2017,Rossi2021} achieving dramatic reductions by exploiting the high skewness of the sample set.

A summary of design parameters of the considered multi-stage classifier principles is given in Table \ref{tab:s2_design_space}. Stage specific parameters are indicated via `$\times$' while parameters, which are globally fixed for the complete cascaded classifier are indicated via `-'.
For example the `$\times$' in the row `number of labels' stipulates that the number of labels can vary from stage to stage, i.e.,  increase towards the last stage.
\begin{table}[htbp]
\vspace{-0.2cm}
	\caption{Stage Specific versus Globally Fixed Design Parameters for Cascaded Classifiers}
	\label{tab:s2_design_space}
	\centering
	\begin{tabular}{|c|c|c|c|c|c|}
		\hline
		\textbf{Design Parameters}&\textbf{\cite{Goetschalckx2018}}&\textbf{\cite{Cocana2017}}&\textbf{\cite{Venkataramani2015}}& \textbf{\cite{Kouris2018}} & \textbf{\textit{ours}} \\
		\hline
		Threshold & $\times$& - & $\times$ &  -  & $\times$ \\
		\hline
		Confidence Metric & - & - & - &  -  & $\times$ \\
		\hline
		Number of Labels &  $\times$ & $\times$ & - & -  &  -\\
		\hline
		Classifier Algorithm & - & - & $\times$ & -   & $\times$ \\
		\hline
	\end{tabular}
\end{table}

As earlier mentioned, the approaches in Fig.\ref{fig:s2_sota_cascaded_classifiers}a) and Fig.\ref{fig:s2_sota_cascaded_classifiers}c) are beneficial for skewed datasets, while the approach in Fig.\ref{fig:s2_sota_cascaded_classifiers}b) is best suited for unskewed datasets, like MNIST, CIFAR\thinspace{10} or ImageNet. 
These datasets are commonly used in benchmarking of machine learning architectures. Hence, the state-of-the-art comparison of cascaded classifiers in Table \ref{tab:sota} focuses on these. 
The key criterion for the comparison is the cost reduction achieved with the cascaded classifier either by upholding the accuracy of the reference classifier or by tolerating 99\% preservation of the reference accuracy.
\begin{table}[htbp]
\vspace{-0.2cm}
	\caption{State of the Art Cascaded Classifiers for Unskewed Datasets}
	\label{tab:sota}
	\centering
		\begin{tabular}{|c|c|c|c|c|c|c|}
		    \hline
		    \textbf{Paper}&\textbf{\cite{Venkataramani2015}}& \textbf{\cite{Kouris2018}}&  \textbf{ours} & \textbf{ours}\\
            \hline
            {Accuracy}& Top-1 & Top-5  & Top-1 & Top-1\\
            \hline
            {Benchmark} & MNIST & ImageNet  & MNIST & CIFAR\thinspace{10}\\
            \hline
            {Cost Function} & ${\text{Energy}}_{\text{norm}}$ & GOps/s & MOps & GOps\\
            \hline
            {Classifier}&  SVM & CNN & CNN & CNN\\
            \hline\hline
             ${\text{acc}_\text{ref}}^{\thinspace{a}}$ &  - & 87\%  & 99.45\% & 94.66\%\\
            \hline
            \begin{tabular}{@{}c@{}}Cost   @ acc$_\text{ref}$\end{tabular} & - & 299.07 & 30.94 & 3.48\\
            \hline\hline
            \begin{tabular}{@{}c@{}c@{}}Cost Reduction \\Cascaded Classifier\\ @ $\text{acc}_\text{ref}^{\thinspace{b}}$\end{tabular} & 1.05$\times$& 0.7$\times$ &5.71$\times$ & 1.32$\times$\\
            \hline
            \begin{tabular}{@{}c@{}c@{}}Cost Reduction \\Cascaded Classifier\\ @ 99\%$\thinspace{\text{acc}_\text{ref}}^{\thinspace{c}}$\end{tabular} & 2.85$\times$ & 1.45$\times$ & 263.17$\times$ & 2.55$\times$ \\
            \hline
      \end{tabular}
   \begin{tablenotes}
   	\centering
     \item[a]$^a$ reference accuracy: accuracy of stand alone final stage classifier
     \item[b]$^b$ achieves same accuracy as reference accuracy
     \item[c]$^c$ achieves 99\% of reference accuracy
   \end{tablenotes}
\end{table}

\section{Methodology}
\subsection{Architecture \& Quantitative Optimization}

This work is based on the general architecture shown in Fig.\,\ref{fig:s2_sota_cascaded_classifiers}b). 
The presented methodology can also be used as in-place replacement of any single stage in Fig. \ref{fig:s2_sota_cascaded_classifiers}c) in order to combine the benefits of both cascaded classifier schemes. 

Our methodology makes use of readily available state-of-the-art
classifier architectures reusing the available outputs to derive a confidence metric.
With this approach, we are generalizing from the ensemble of one-class-expert classifiers as in \cite{Venkataramani2015} to a set of existing, and upcoming, machine learning classifiers freely chosen by their key performance indicators, e.g. cost and accuracy. 

Given  the complexity  of  comparing  different  solutions, cost is assumed to scale proportional to the number of Multiply-and-Accumulate (MAC) operations per inference and according to the quantization. 
Starting with a 32-bit floating point (fp32) reference implementation, scaling factors for quantized 32-bit fixed point (fx32) and binary (bin) operations have to be identified.
To scale to fx32, we use a factor of $\alpha_{\text{fp32}\rightarrow \text{fx32}}=0.7$ computed as ratio of sums of energy numbers of addition and multiplication given in \cite{Horowitz2014}.
Scaling to binary representation according to \cite{Stadtmann2020} would result in a factor of $\alpha_{\text{fx32}\rightarrow \text{bin}}>2000$.
Well aware of the crudeness of this assessment, we decided to adopt as upper bound a rather conservative scaling factor of $\alpha_{\text{fp32}\rightarrow \text{bin}}=10^3$.
The above assumption implies that the average cost for memory accesses and data transfers scales accordingly to the MAC operations.
Focusing on relative gains between different architectures, this is considered a sufficiently accurate first order approximation of the involved cost in terms of average processing time or energy per sample.

Cost reduction is obtained if the pass-on-probability $\rho_{i}$ of a sample in a classifier stage $i$ is selected as to meet inequality $C_{i} + \rho_{i} \cdot C_{i-1} \leq C_{i-1}$ or $\rho_{i} \leq 1 - C_{i}/C_{i-1}.$
So, the cost introduced by an additional stage is overcompensated by its savings.
At the same time, $\rho_{i}$ needs to be adjusted individually per classifier stage to achieve the targeted accuracy while minimizing the cost. 
If all samples would pass through all $N$ stages, total cost would increase to $C_{\mbox{tot}}=\Sigma_{i=0}^NC_i$.

\subsection{Vehicle}
To elaborate the methodology, we use the basic MNIST digit recognition example.
The study features three different Artificial Neural Network (ANN) architectures trained to classify MNIST samples.
For this case of balanced datasets, the quality of an ANN is commonly measured in terms of accuracy defined by the number of correctly identified samples divided by the number of all samples.
As a baseline, we use LeNet5 \cite{Lecun1998} with small modifications in order to achieve a competitive accuracy of 99.45\% (cf.\,Table \ref{tab:s3_kpi_mnist}).
The second ANN is a Multi-Layer Perceptron (MLP) with a single hidden layer (512 neurons) further denoted as FC3.
The smallest ANN is a result of binarization (cf.\,\cite{Courbariaux2016}) of the FC3 denoted as FC3\_bin. 
With increasing number of classifiers, the savings diminish \cite{Goetschalckx2018}.
Hence, in this study we limit the maximum number of stages to three.
A summary of the key performance indicators of the three ANNs is given in Table \ref{tab:s3_kpi_mnist}. 
The hyperparameters for training the ANNs are found using grid search, since the iterations on small datasets and networks are relatively short. For larger datasets, we would recommend optimization tools (such as Hyperopt \cite{bergstra2015}) to converge to a good solution faster. This was not performed within the scope of this work.
The ANNs in this paper are state-of-the-art architectures for hardware implementations, which achieve state-of-the-art accuracies without extensive data augmentation.

\begin{table}[htbp]
\vspace{-0.3cm}
	\caption{Key Performance Indicators of MNIST Classifiers}
	\centering
	\begin{tabular}{|c|c|c|c|}
		\hline
		\textbf{ANN}&\textbf{Benchmark}&\textbf{Cost}& \textbf{Accuracy} \\
		\textbf{}&\textbf{}&\textbf{\textit{C}}& \textbf{\%} \\
		\hline
		FC3\_bin & MNIST & 410 & 95.29  \\
		\hline
		FC3 & MNIST & 410\thinspace k & 98.43 \\
		\hline
		LeNet5 & MNIST & 31\thinspace M & 99.45 \\
		\hline
	\end{tabular}
	\label{tab:s3_kpi_mnist}
	\vspace{-0.1cm}
\end{table}

\subsection{Analyses of Pass-on Criteria}
\subsubsection{Confidence Metric}
An essential step to reuse existing single-stage classifiers in a cascade is the definition of a pass-on criterion, i.\,e. a way to reject decisions made by a classifier at a specific stage for further analyses in a succeeding classifier stage.
The pass-on criterion consists of a specific confidence metric and its corresponding threshold level.
Applying the softmax function (cf. Eq.\,\ref{eq:s3_softmax}) to the raw output vector $\mathbf{x}$ of the last an ANN layer results in the vector of probabilities $\mathbf{x}^\text{s}$
indicating the probability of a sample to belong to class $i$, with $L$ being the total number of labels.

\begin{equation}
\label{eq:s3_softmax}
x^{\text{s}}_{i} = \frac{e^{x_i}}{\sum_{j=1}^{L}e^{x_j}}
\end{equation}

To provide a less computationally complex alternative to the softmax function, we propose a linear normalization of the output vector $\mathbf{x}$
by means of Eq.\,\ref{eq:s3_linear}.
This normalization guarantees, that the resulting vector $\mathbf{x}^\text{l}$ has the
properties of 
a probability distribution, i.e.,  $x_i^\text{l} \in [0,1]$ and $\sum_{i=1}^{L} x^\text{l}_i = 1$.

\begin{equation}
\label{eq:s3_linear}
x^{\text{l}}_{i} = \frac{(x_i - x_{\text{min}})}{\sum_{j=1}^{L}(x_j - x_{\text{min}})}
\end{equation}

Both normalization schemes enable the utilization of statistical confidence metrics.
The most used confidence metric is the absolute maximum value of the normalized output vector $\mathbf{x}^\text{s}$ or $\mathbf{x}^\text{l}$ here called ({\sc abs}). 
Usually, this value is compared to a given threshold.
We extend the selection of confidence metrics, which are applied to the normalized output vector, to the following selection:

\begin{itemize}
	\item `Best guess Versus the Second Best guess' ({\sc bvsb}) 
	\item Variance ({\sc var})
	\item Entropy ({\sc ent})
	\item Kullback-Leibler Divergence ({\sc kl\_div})
	\item Kurtosis ({\sc kurt})
\end{itemize}

The {\sc bvsb} computes the absolute difference between the top-two results \cite{Joshi2009}. Its generalized form \cite{Kouris2018} is not considered further, as we saw no relevant improvement in our benchmarks.
This paper is the first to the authors knowledge, which introduces the {\sc kurt} and {\sc kl\_div} as confidence metrics for cascaded classifier.
For all statistical metrics, the elements of the output vector are treated as samples from a probability density function.
{\sc kurt}, defined as the fourth standardized moment, is linearly normalized between the worst case (uniform distribution) and the best-case (delta distribution).
Regarding {\sc kl\_div}, the reference distribution is defined as a delta distribution, representing full confidence into one label.

\subsubsection{Pass-On-Probability}
\begin{figure}[htbp]
	\centering
	\includegraphics[scale=0.77]{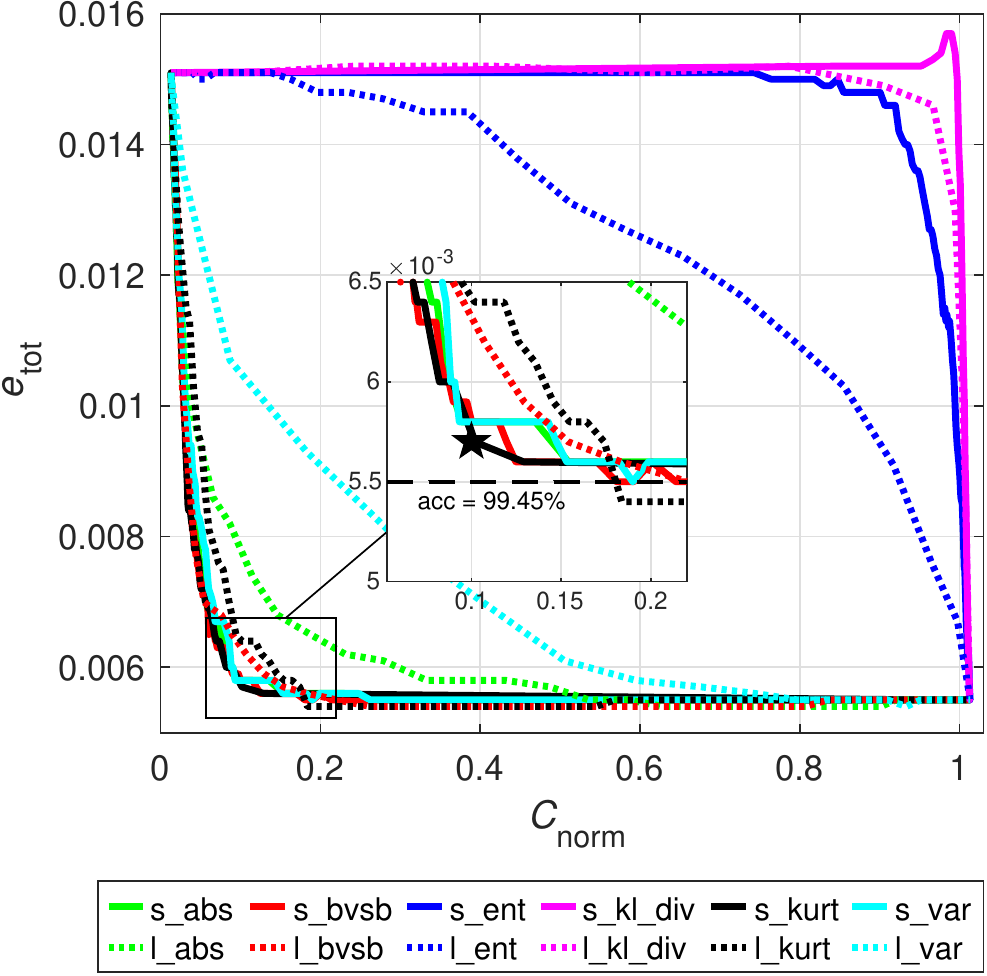}
	\caption{Pareto analysis of pass-on criteria for softmax `s' and linear `l' output of 2-stage classifier using FC3 as initial stage followed by reference LeNet5.}
	\label{fig:s3_fc3_ppplot}
\end{figure}

Based on the previously introduced confidence metrics, a pass-on criterion can be set to identify otherwise incorrectly classified samples, that should be passed on to the next classification stage with a more capable classifier.
To benchmark the different confidence metrics introduced in the previous chapter, we set up a 2-stage cascaded classifier but focus on the behavior in the first stage.
For the first stage, either one of FC3 or FC3\_bin is used. 
In this case, LeNet5 acts as last stage (reference) making any final decision on unclassified samples.
Sweeping the threshold for various confidence metrics, the accuracy of the first and last stage will be reached at thresholds $th=0$ and $th=1$, respectively.
At the same time, the cost scales from the first stage's to the sum of both stages.
An example of such sweep is shown in Fig.\thinspace\ref{fig:s3_fc3_ppplot} for the FC3. It visualizes the total error ($e_{\text{tot}}$ = 1 - accuracy) of the cascaded classifier vs. the normalized cost $C_\text{norm}$, i.e.,  the total cost divided by the cost of the reference classifier $C_\text{norm} = C_{\text{tot}} / C_0$. 

Considering confidence metrics entropy {\sc ent} or Kullback-Leibler divergence {\sc kl\_div}, the error of the cascaded classifier only drops for very large thresholds, i.\,e., when the majority of samples are passed on to the reference classifier. 
This is true for any of the two first stage classifiers and appears largely independent whether the confidence metric is based on $\mathbf{x}^\text{l}$ or $\mathbf{x}^\text{s}$.
Hence, these confidence metrics are excluded from further considerations.

Between the remaining four confidence metrics, there is a less apparent difference. 
The inset in the figure shows a zoom at the knee point of the curves.
Out of the remaining confidence metrics, the ones based on {\sc bvsb} and {\sc kurt} have the tendency of showing better results, i.\,e., they achieve a better trade-off in terms of error vs.\ cost as they are closer to the lower left corner of the figure.
For instance the cost reduces by 90\% while tolerating an increase of 0.02\% in the classification error for {\sc kurt} on $\mathbf{x}^{\text{s}}$ (black star in Fig.\thinspace\ref{fig:s3_fc3_ppplot}).
In the MNIST case, this corresponds to another two (out of 10k) misclassified samples.
The combination of both classifier stages shows for {\sc kurt} on $\mathbf{x}^{\text{s}}$ a slight improvement in classification accuracy compared to the stand alone final stage classifier (cf.\,inset in Fig.\thinspace\ref{fig:s3_fc3_ppplot}).
This leads to the conclusion that for selected samples the first stage classifier provides correct labeling, whereas in a following stage these samples experience a small probability of being falsely classified.

This experiment shows that the simple confidence metric, maximum value {\sc abs}, already provides a good selection of relevant samples, but can be improved by integrating more information available from the output vector. 
As much as the computation of the presented confidence metrics might not be relevant in terms of total cost, they add to latency especially when the employed transcendental functions are computed iteratively on hardware.
In the end, as the difference between the two best confidence metrics ({\sc kurt} and {\sc bvsb}) is rather small (being based on linear or softmax output vectors), a final selection will be based on the CIFAR\thinspace{10} validation vehicle as discussed in Section \ref{sec:casestudy}.

In conclusion, a threshold applied to an appropriate confidence metric can identify correctly classified samples so that only a minimal number of samples has to be passed on to the following stage.

\subsection{Generation of Multi-Stage Classifiers}
\label{ssec:mnist_2stage}

Based on the earlier assessment, a multi-stage classifier has to be created in a trial-and-error fashion iterating through various sequences of ANNs.
Thereby, the accuracy and cost should increase going deeper in the hierarchy.
In a first step, the earlier introduced LeNet5 is combined with either FC3 or FC3\_bin to result in a 2-stage cascade.
As before the threshold is swept from 0 to 1 to determine the achievable trade-offs between accuracy and cost as a function of realized pass-on-probability 
$\rho_i \in [0,1]$.

\begin{figure}[htbp]
	\centering
	\includegraphics[scale=0.88]{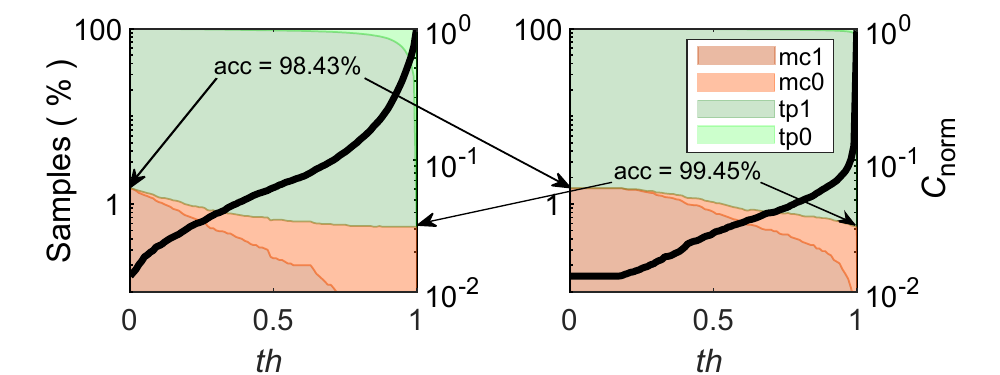}
	\caption{In-depth analysis of true positives and misclassified samples of the initial classifier FC3 and the reference classifier LeNet5 for the confidence metrics {\sc bvsb} (left) and {\sc kurt} (right) with a softmax activation.}
	\label{fig:s3_fc3_tpmcplot}
	\vspace{0.2cm}
\end{figure}
Firstly, the resulting classifier behavior is visualized in Fig.\thinspace\ref{fig:s3_fc3_tpmcplot} for the case of the FC3-LeNet5 combination.
As expected, the percentage of misclassified samples (redish coloring) diminishes for higher thresholds, i.\,e., when passing more samples to the final stage.
Also visible is the trend of decreasing number of misclassifications in stage 1 and their related increase in stage 0. 
However, as stage 0 is a superior classifier the overall error count is reduced.
This trend is shown for the two overall best performing confidence metrics {\sc bvsb} and {\sc kurt} for $\mathbf{x}^\text{s}$ on the left- and right-hand-side, respectively.
As much as the reduction in error occurs for lower thresholds in the case of {\sc bvsb}, it can be attributed to an earlier increase in $\rho$. 
The cost level (solid black line) appears almost identical for same accuracy levels.
However, for equidistant sampling of the threshold $th$ the pass-on-probability $\rho$ is more evenly distributed in the case of {\sc bvsb}, which is not the case for {\sc kurt}.
We conclude, that {\sc bvsb} offers a finer tuning range.

\begin{figure}[htbp]
	\centering
	\includegraphics[scale=0.77]{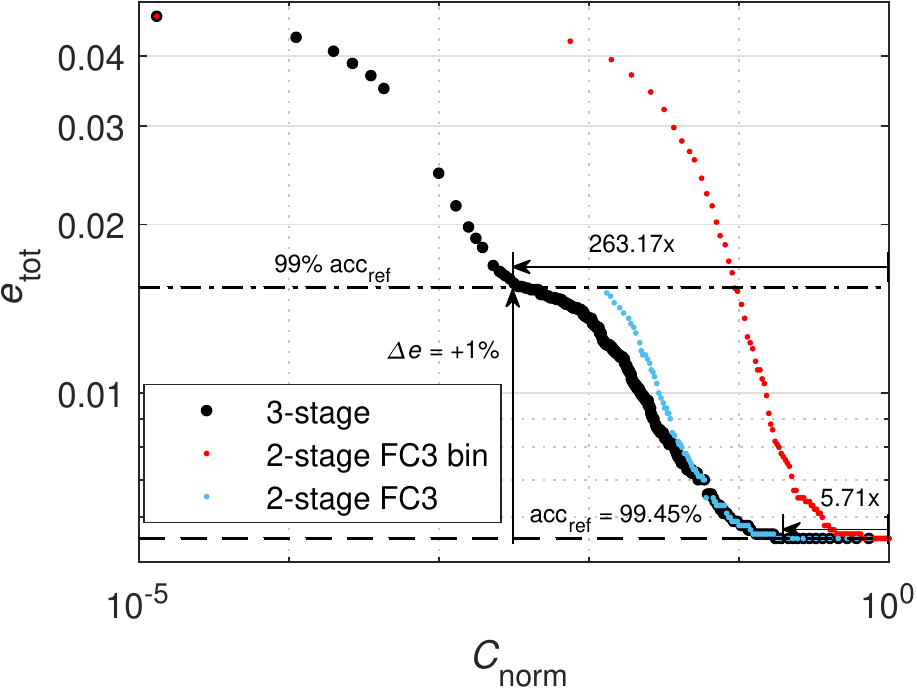}
	\caption{Pareto plot comparing the 2-stage classifier (red and blue) with the 3-staged classifier in use case MNIST (black dots enlarged for better legibility).}
	\label{fig:s3_3stage_mnist_logppplot}
	\vspace{0.2cm}
\end{figure}

In a second step, all three ANNs are sequentially operated resulting in a 3-stage classifier.
Hereby, the classifiers are staged sorted by their increasing accuracy and cost.
The thresholding is performed using the softmax activation function and {\sc bvsb} as a confidence metric. 
For the earlier cases of 2-stage classifiers basically all values of the threshold resulted in a potential operating point.
In the case of multiple stages the sweep results in a large collection of design points. 
Hence, Pareto-optimization is applied to the results of a full-factorial analysis across a fine granular sweep of both classifier stages.
The resulting Pareto-optimal points are dominant combinations of error and cost (cf.\,Fig.\,\ref{fig:s3_3stage_mnist_logppplot}).
Consequently, these combinations achieve the lowest error for a certain cost.
Analogous to the previous experiments, the cost is normalized against the final stage $C_0$.
For reference, the previous 2-stage classifiers FC3\_bin-LeNet5 and FC3-LeNet5 are indicated in red and blue, respectively.
The operating points of the 3-stage classifier are shown as black dots.
As much as the 2-stage variants provide a trade-off between the extreme points, they lack the much wider, useful scaling range offered by the 3-stage implementation.
In all cases, the 3-stage classifier achieves a significantly better operating point across the 5 orders of cost - offering graceful degradation in accuracy as a function of available cost.

\section{Case Study: CIFAR\thinspace{10}}
\label{sec:casestudy}

To validate the methodology developed using the basic MNIST example, it is now applied to the case of CIFAR\thinspace{10} classification.
As before, three ANNs are selected, each capable to perform the classification task stand-alone.
Please note, the training of the ANNs for CIFAR\thinspace{10} (summarized in Table \ref{tab:s4_kpi_cifar}
\footnote{Note that the width and depth of the ANNs used for this case study are adjusted for the CIFAR\thinspace10 dataset.}
utilized only basic data augmentation methods like padding and shifting from the center.
Further improvement in classification accuracy is reached with hyperparameter optimization and heavy data augmentation (cf.\,\cite{Lin2015}). 
Since DenseNet\cite{Huang2016} achieves the best accuracy (at highest cost), it is selected as reference classifier, i.\,e.\ as last stage in the cascade and reference of cost.

\begin{table}[htbp]
	\vspace{-0.2cm}
	\caption{Key Performance Indicators of CIFAR\thinspace{10} Classifiers}
	\centering
	\begin{tabular}{|c|c|c|c|}
		\hline
		\textbf{ANN}&\textbf{Benchmark}&\textbf{Cost}& \textbf{Accuracy} \\
		\textbf{}&\textbf{}&\textbf{\textit{C}}& \textbf{\%} \\
		\hline
		LeNet5 & CIFAR\thinspace{10} & 32.5\thinspace M & 83.55 \\
		\hline
		VGG7  & CIFAR\thinspace{10} & 2.6\thinspace G & 92.78 \\
		\hline
		DenseNet  & CIFAR\thinspace{10} & 3.5\thinspace G & 94.66 \\
		\hline
	\end{tabular}
	\label{tab:s4_kpi_cifar}
	\vspace{-0.1cm}
\end{table}

As for MNIST, an optimal confidence metric has to be selected for CIFAR\thinspace{10}. There are two candidate confidence metrics for MNIST, which hold promising results, but in the CIFAR\thinspace{10} case, the {\sc bvsb} shows a clear edge over {\sc kurt}. Hence, this confidence metric is used in the following experiment. Please note, as for the MINST example, also CIFAR\thinspace{10} shows a slight improvement in accuracy for intermediate threshold levels.
Bear in mind, that all points with $C_{\text{norm}} \ge 1$ are indicating higher cost than a stand-alone reference classifier.

\begin{figure}[htbp]
	\centering
	\includegraphics[scale=0.77]{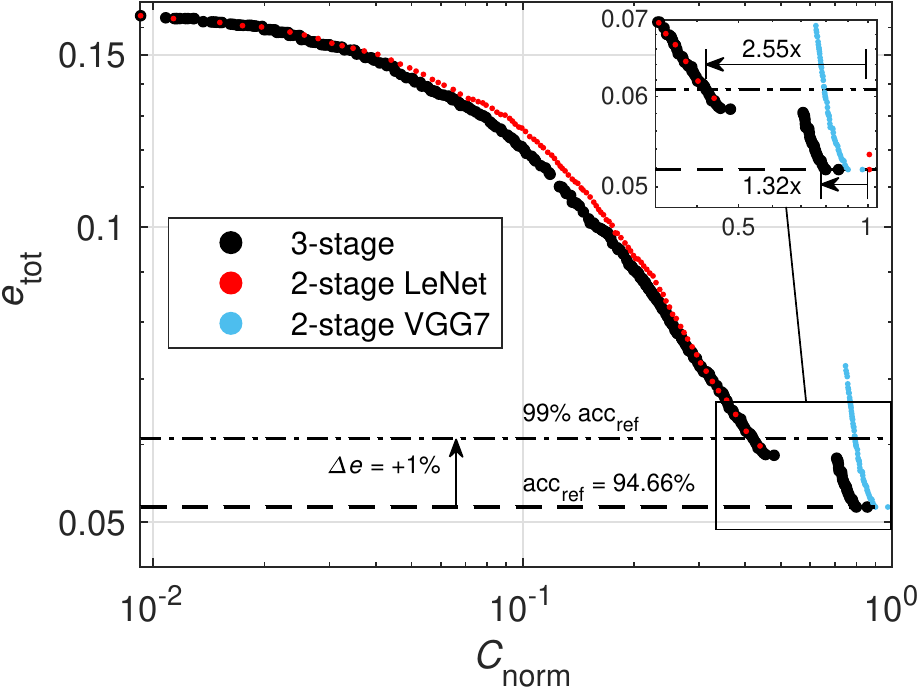}
	\caption{Pareto plot comparing the 2-staged classifier (red and blue) with the 3-staged classifier (black dots enlarged for better legibility) in use case CIFAR\thinspace{10}.}
	\label{fig:s4_3stage_cifar_lenet_vgg7_dense_softmax_diff_logppplot}
\end{figure}

Fig.\thinspace\ref{fig:s4_3stage_cifar_lenet_vgg7_dense_softmax_diff_logppplot} depicts the Pareto plot of the 3-staged classifier. 
The gap in the Pareto plot indicates that there is no reduction in classification error for this cost interval.
With the chosen classifiers, it is possible to reduce the cost over two orders of magnitude.
A wider range, as obtained for the 3-stage MNIST classifier, would be possible using an additional ANN of lower cost and reasonable accuracy.
While the accuracy of the reference classifier is maintained, the cascaded classifier achieves a cost reduction of 1.32$\times$ by using low-cost classification of easy samples.
The cost reduction becomes 2.55$\times$ maintaining 99\% of the reference accuracy.

\section{Conclusion}

Ever more powerful machine learning algorithms surpass human performance, yet are prohibitive for embedded devices due to their high computational complexity. Combining classifiers of varying accuracy-to-cost ratios in a cascade, as presented in this paper, provides cost reduction, while preserving top accuracy. Alternatively, graceful degradation is enabled to provide dramatic cost reduction, including throughput and energy,
with bounded drop in accuracy. 
The realization of such cascaded classifier can be reproduced in the presented structured methodology, which makes use of a confidence metric requiring only basic mathematical operations. Pareto-optimization provides optimal settings for the threshold level to adapt classification on embedded devices according to the available energy budget. 
The presented methodology is derived using MNIST and validated with CIFAR\thinspace10. This work achieves with
a three stage classifier for MNIST a cost reduction of 5.71$\times$ at 99.45\% accuracy and remarkable 263.17$\times$ at 98.46\%. In the case of CIFAR\thinspace10, the reduction is 1.32$\times$ at 94.66\%, and 2.55$\times$ at 93.7\%. 

\subsubsection{Acknowledgment.}
This work was partially funded by the German BMBF project NEUROTEC under grant no. 16ES1134.

 \bibliographystyle{splncs04}
 \bibliography{lodbib}

\end{document}